\documentclass[runningheads]{llncs}
\usepackage[T1]{fontenc}
\usepackage{graphicx}
\usepackage[table,xcdraw]{xcolor}
\usepackage[noend]{algorithmic}
\usepackage{algorithm}
\usepackage[normalem]{ulem}
\useunder{\uline}{\ul}{}
\usepackage{layouts}
\usepackage{comment}
\usepackage{multirow}
\usepackage{paralist}
\usepackage{url}
\usepackage{microtype}
\usepackage{calc}
\usepackage{booktabs}

\makeatletter
\newcommand\thefontsize{The current font size is: \f@size pt}
\newcommand{\showFont}{encoding: \f@encoding{},
  family: \f@family{},
}
\newcommand\x{5px}
\newcommand\y{0px}
\DeclareRobustCommand{\saclass}{%
  \begingroup\normalfont
  \includegraphics[height=\fontcharht\font`\B+\y, width=\fontcharht\font`\B+\x]{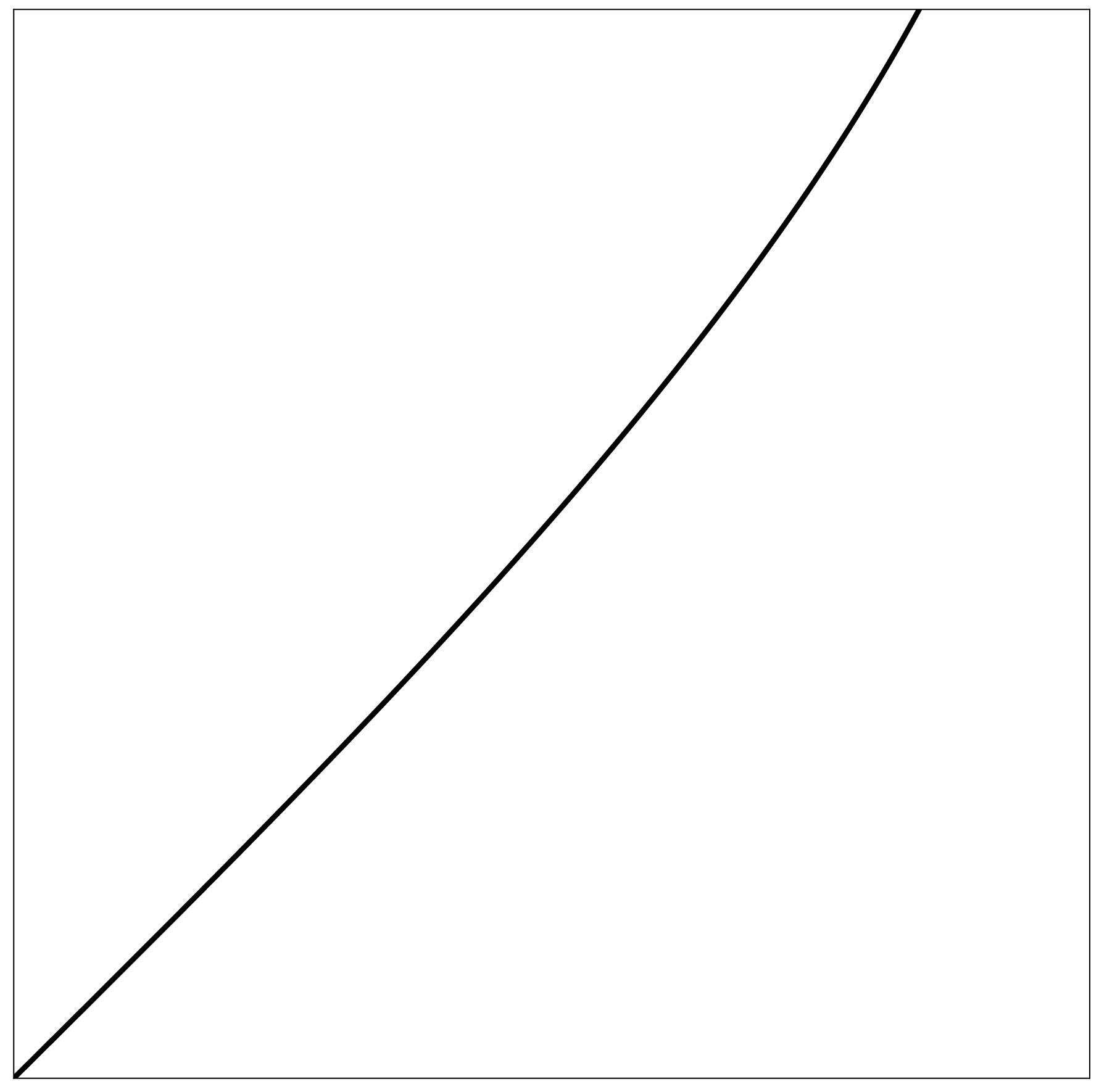}%
  \endgroup
}
\DeclareRobustCommand{\sbclass}{%
  \begingroup\normalfont
  \includegraphics[height=\fontcharht\font`\B+\y, width=\fontcharht\font`\B+\x]{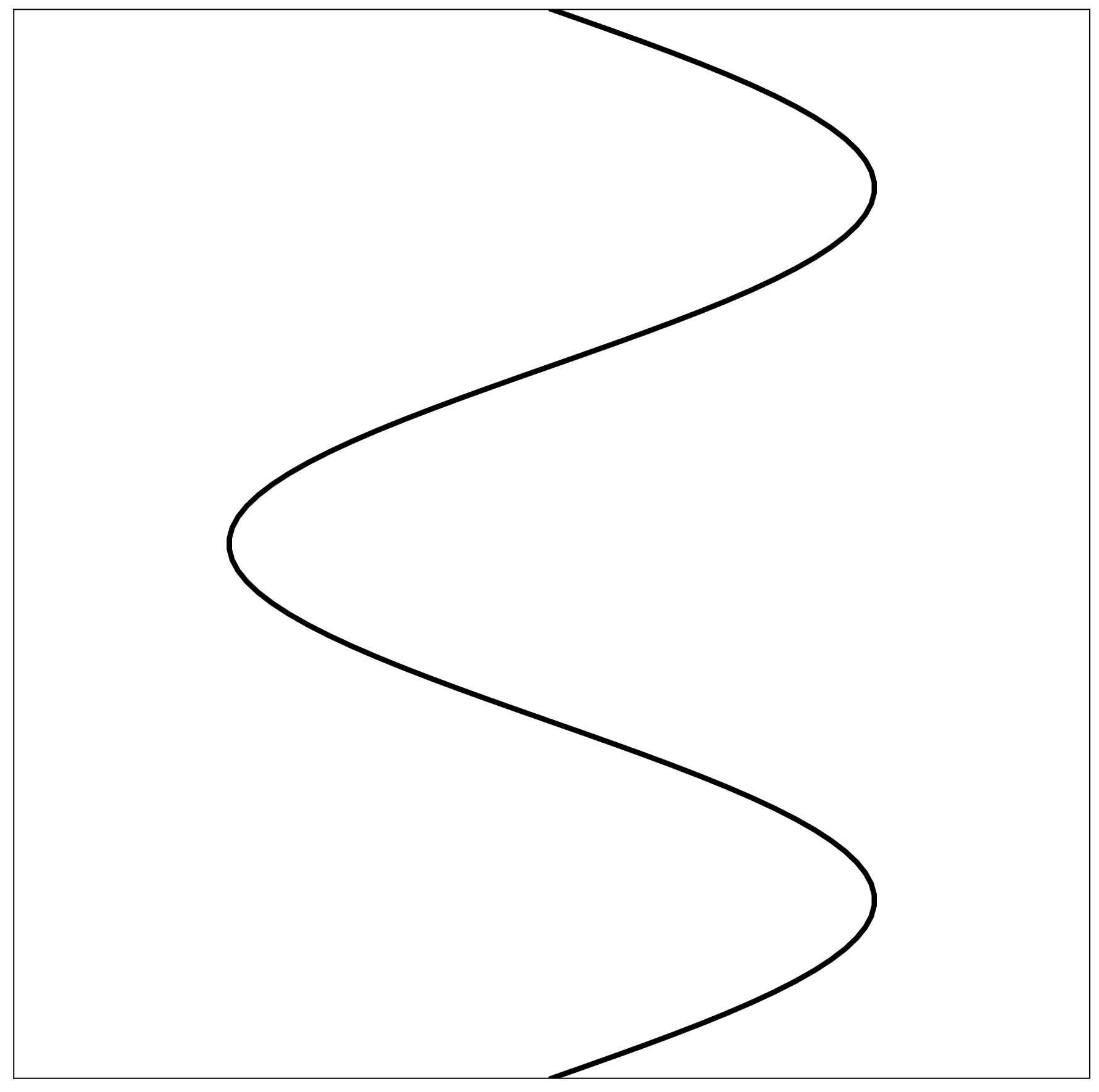}%
  \endgroup
}
\DeclareRobustCommand{\saplus}{%
  \begingroup\normalfont
  \includegraphics[height=\fontcharht\font`\B+\y, width=\fontcharht\font`\B+\x]{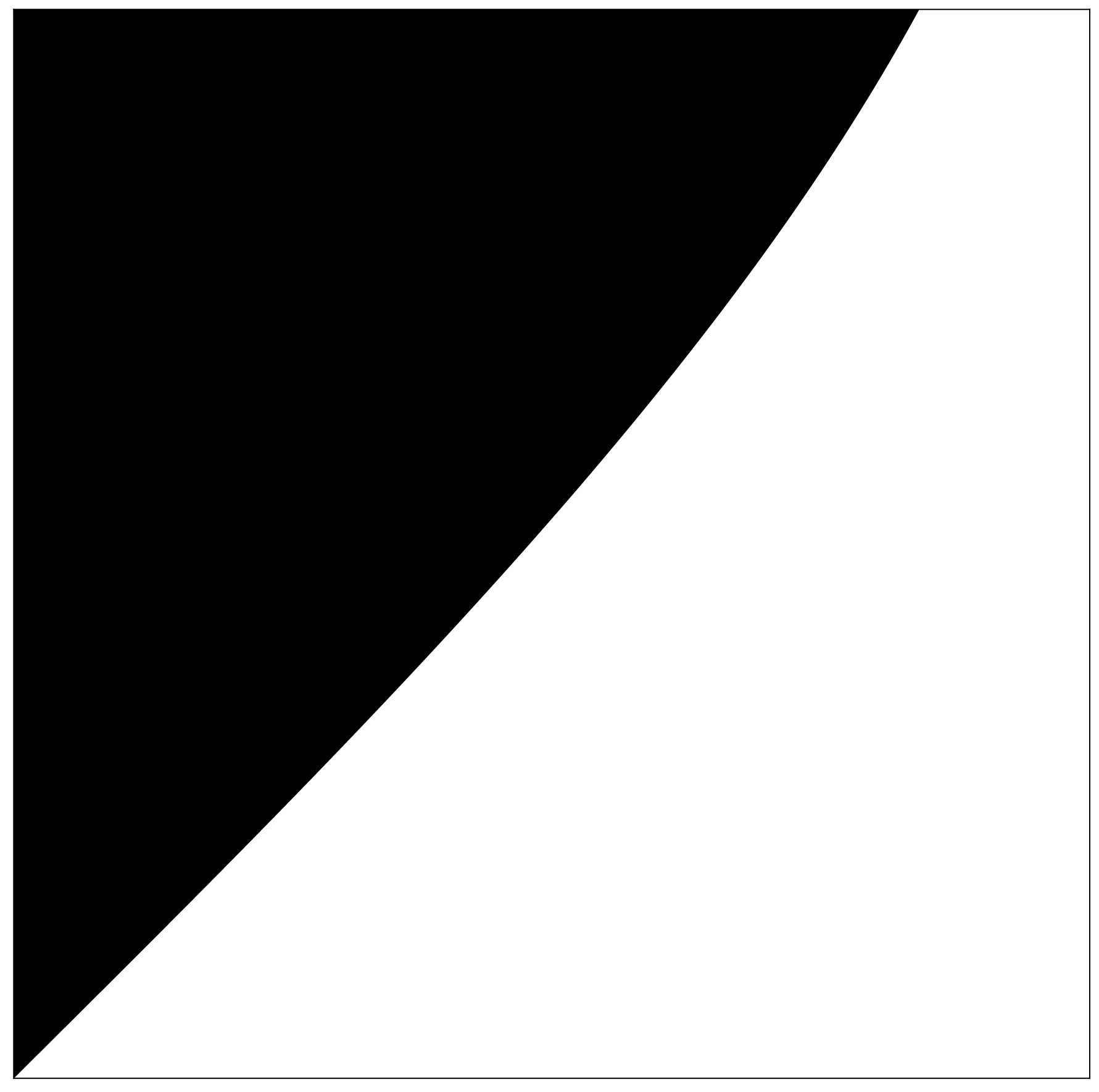}%
  \endgroup
}
\DeclareRobustCommand{\saminus}{%
  \begingroup\normalfont
  \includegraphics[height=\fontcharht\font`\B+\y, width=\fontcharht\font`\B+\x]{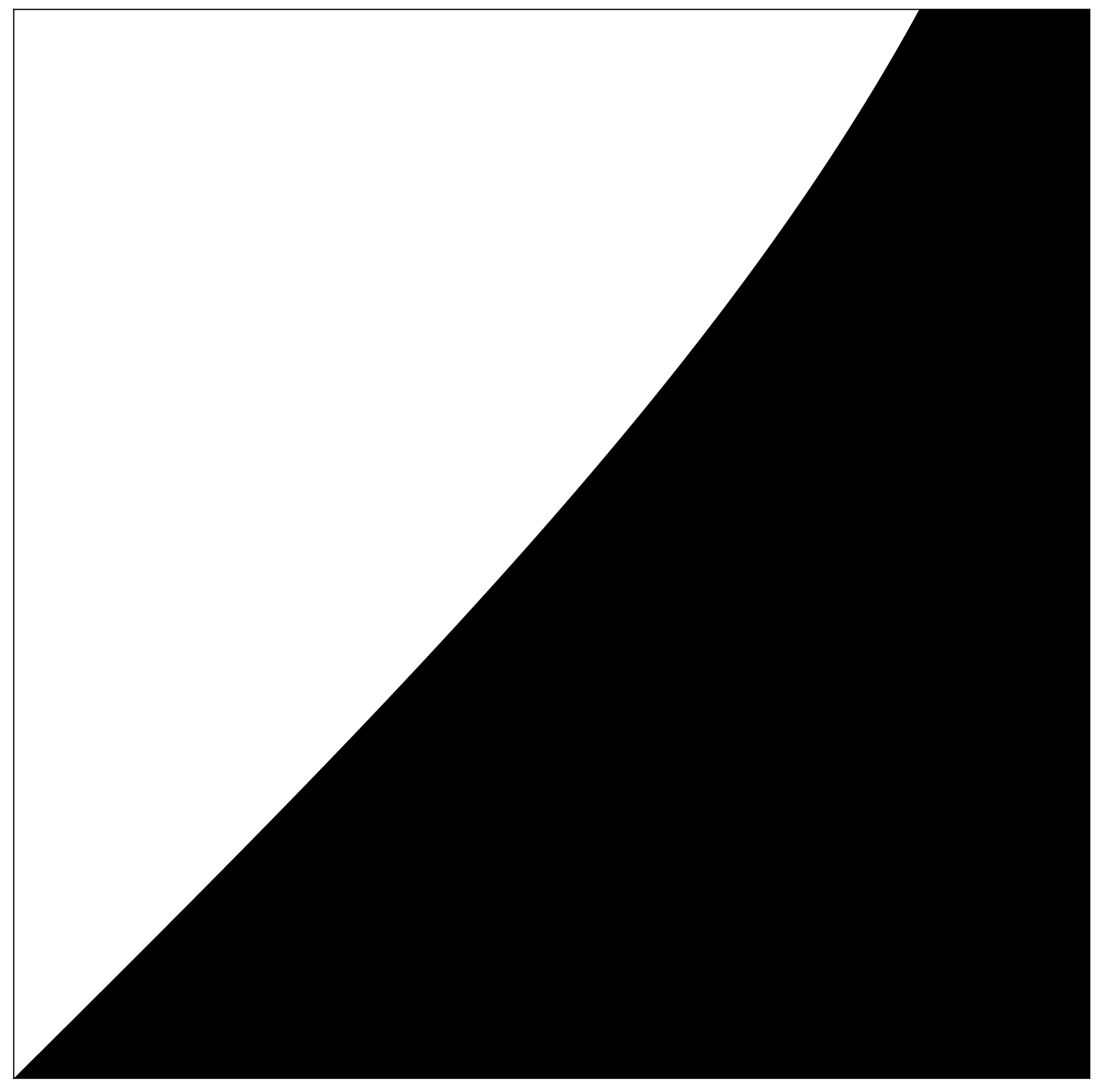}%
  \endgroup
}
\DeclareRobustCommand{\sbplus}{%
  \begingroup\normalfont
  \includegraphics[height=\fontcharht\font`\B+\x, width=\fontcharht\font`\B+\y, angle=90]{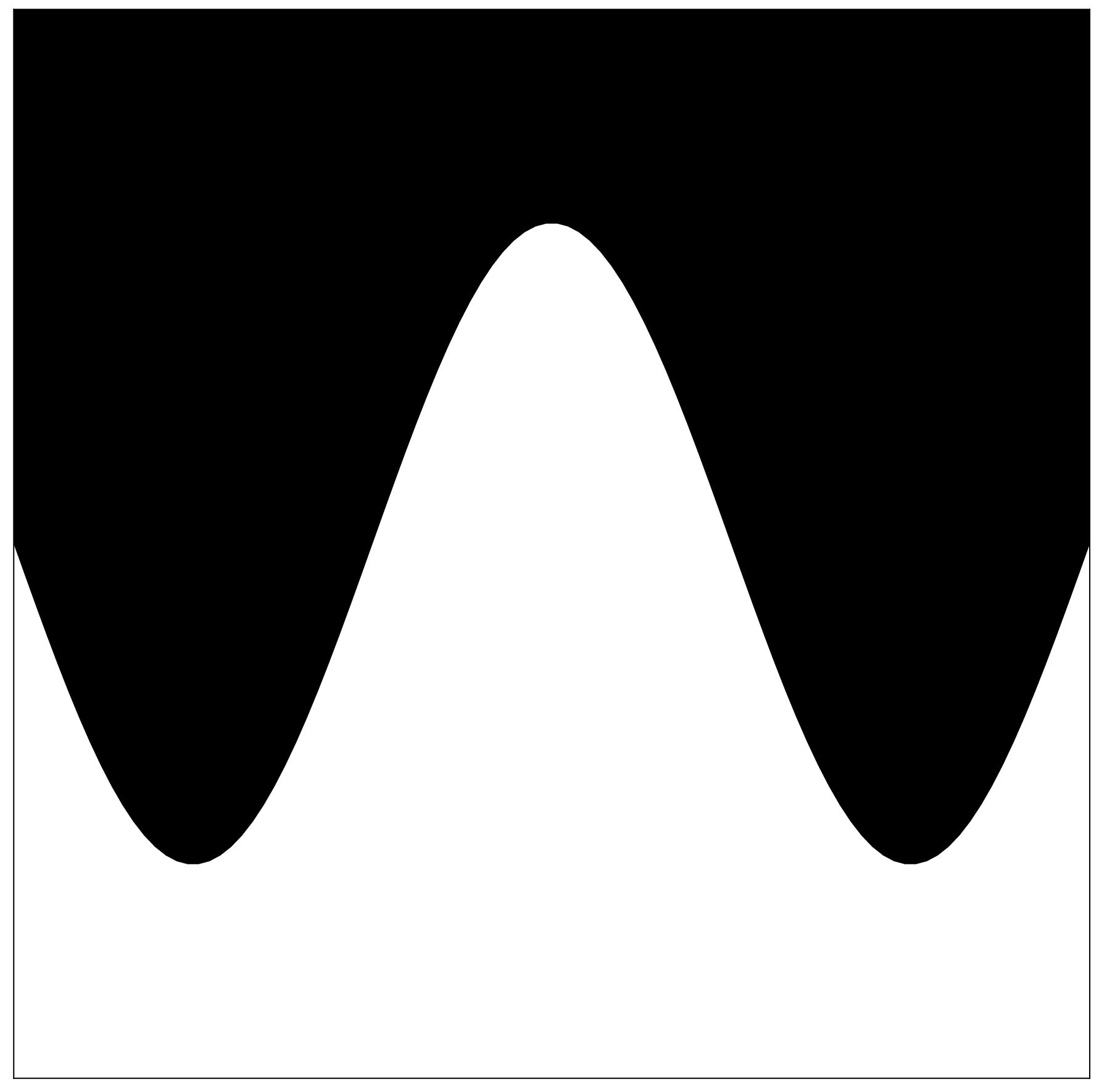}%
  \endgroup
}
\DeclareRobustCommand{\sbminus}{%
  \begingroup\normalfont
  \includegraphics[height=\fontcharht\font`\B+\x, width=\fontcharht\font`\B+\y, angle=90]{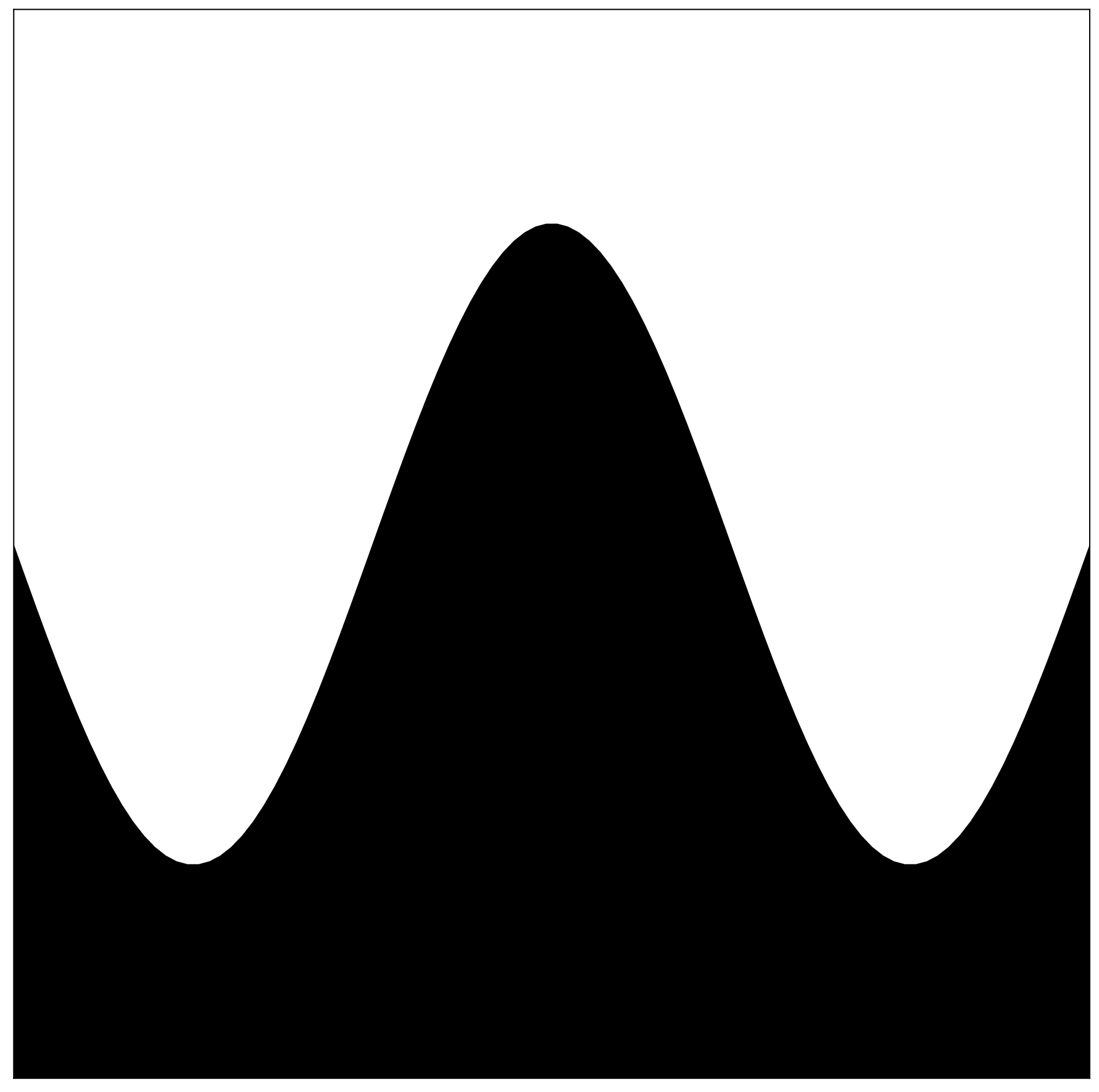}%
  \endgroup
}

\newcommand\blfootnote[1]{%
  \begingroup
  \renewcommand\thefootnote{}\footnote{#1}%
  \addtocounter{footnote}{-1}%
  \endgroup
}

\makeatother
\begin{document}
\title{cPNN: Continuous Progressive Neural Networks for Evolving Streaming Time Series}

\author{Federico Giannini\inst{1}\orcidID{0000-0002-4210-6271} \and Giacomo Ziffer\inst{1}\orcidID{0000-0002-2768-3580} \and Emanuele Della Valle\inst{1}\orcidID{0000-0002-5176-5885}}
\authorrunning{F. Giannini et al.}
%
\institute{DEIB, Politecnico di Milano, Milano, Italy 
\email{\{federico.giannini,giacomo.ziffer,emanuele.dellavalle\}@polimi.it}}

\titlerunning{cPNN: Continuous PNNs for Evolving Streaming Time Series}
\maketitle              
\blfootnote{This is the author’s accepted version of the paper:\\
Giannini, F., Ziffer, G., Della Valle, E. (2023). \textbf{cPNN: Continuous Progressive Neural Networks for Evolving Streaming Time Series}. In: PAKDD 2023, Lecture Notes in Computer Science.
The final publication is available at Springer via \url{https://doi.org/10.1007/978-3-031-33383-5_26}\\
\copyright Springer Nature 2023.}

\begin{abstract}
Dealing with an unbounded data stream involves overcoming the assumption that data is identically distributed and independent. A data stream can, in fact, exhibit temporal dependencies (i.e., be a time series), and data can change distribution over time (concept drift). The two problems are deeply discussed, and existing solutions address them separately: a joint solution is absent. In addition, learning multiple concepts implies remembering the past (a.k.a. avoiding catastrophic forgetting in Neural Networks' terminology). This work proposes Continuous Progressive Neural Networks (cPNN), a solution that tames concept drifts, handles temporal dependencies, and bypasses catastrophic forgetting. cPNN is a continuous version of Progressive Neural Networks, a methodology for remembering old concepts and transferring past knowledge to fit the new concepts quickly. We base our method on Recurrent Neural Networks and exploit the Stochastic Gradient Descent applied to data streams with temporal dependencies. Results of an ablation study show a quick adaptation of cPNN to new concepts and robustness to drifts.

\end{abstract}
\section{Introduction}
\label{intro}
In a context where data comes as an unbounded data stream and is continually evolving, we must overcome the central hypothesis of Machine Learning (ML): the assumption according to which data is independent and identically distributed (shortly, i.i.d). It does not hold for any data stream where data could suffer from changes in its distribution (the so-called "concept drift") and shows temporal dependencies. While the literature has deeply investigated the two situations separately, few works deal with the joint problem. The need to find a combined solution is, thus, increasingly emerging. We formalize the mentioned problem by calling it \textbf{Evolving Streaming Time Series (ESTS)}. "Evolving" indicates the possibility of concept drift, while "Streaming" refers to data points arriving continually from an unbounded data stream. We use "Time Series" to stress the presence of temporal dependencies. Working with concept drifts and multiple concepts makes it necessary to consider the well-known stability-plasticity dilemma~\cite{q_stability_plasticity}, according to which too much plasticity results in forgetting past knowledge. This problem is known as \textbf{catastrophic forgetting (CF)}~\cite{q_catastrophic_forgetting}. Too much stability leads, instead, to difficulties in learning new knowledge. 

Among the models for dealing with time series, sequential models based on Recurrent Neural Networks (RNN) are widely used in the literature~\cite{q_deep_learning}. Applying Neural Networks (NN) to the streaming scenario allows it to exploit its learning algorithm's (Stochastic Gradient Descent, SGD) adaptability. In contrast, SGD can suffer when the new concept differs significantly from the previous one, and a NN forgets the last concept when it learns a new one. To resolve these issues, Progressive Neural Networks (PNN)~\cite{q_pnn} consist of NN architectures to jointly remember the previously learned knowledge and use transfer learning to recycle the knowledge gained from old concepts~\cite{q_transfer_learning}. However, this methodology is not meant to deal with an ESTS.

Our work, thus, aims to investigate the following \textbf{research question}: \emph{in the context of an Evolving Streaming Time Series, is there a solution to jointly manage concept drifts, temporal dependencies, and catastrophic forgetting?} In this paper, we positively answer this question by contributing \textbf{Continuous PNN (cPNN)}, a novel continuous version of PNNs that extends them to an ESTS scenario. We first propose a strategy to exploit SGD in a streaming scenario to tame temporal dependencies. Secondly, our approach utilizes PNN-based architectures to efficiently address both concept drifts and CF, using transfer learning to enable rapid adaptation to new concepts while maintaining the predictive ability of previously learned ones. A crucial feature of cPNN is that the architecture can be potentially applied to each type of RNN model. We conduct an ablation study on a binary classification problem during the experiment phase to test cPNN on synthetically generated data streams. We compare cPNN with two ablated architectures: cLSTM and mcLSTM. After a concept drift, cLSTM continues training on the new concept. It, thus, does not avoid CF and is not concept drift aware. mcLSTM avoids CF, but it does not use transfer learning. Temporal dependencies are tamed by using RNN models. Results show that cPNN performs better after concept drifts than ablated architectures.

The rest of the paper is organized as follows. Firstly, Section~\ref{related_works} analyzes the already present ideas in literature. Section~\ref{proposed} exposes our method and contributions. Then, Section~\ref{exp_set} discusses the settings of our experiments, while Section~\ref{results} exhibits the results. Finally, Section \ref{conclusion} discusses conclusions and future works. 

\section{Related Works}
\label{related_works}
\textbf{Continual Learning (CL)} thoroughly investigated methods to learn and avoid CF continually~\cite{q_cl}. The Task Incremental Learning scenario assumes that data is split into batches of samples (named experiences) provided over time. Each of them represents a task. The data distribution and objective function are normally fixed within a task. In this paper, we refer to this scenario whenever we use CL.

\begin{figure}[t!]
\centering
\includegraphics{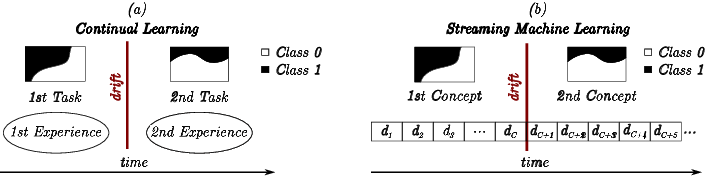}
\caption{Comparison of CL and SML scenarios.} \label{f_cl_sml}
\end{figure}

In this context (shown in Fig.~\ref{f_cl_sml}.a), \textbf{PNNs}~\cite{q_pnn} are NN architectures that use transfer learning to recycle knowledge gained from previous tasks. Furthermore, the parameters associated with the old tasks are frozen to avoid CF. PNNs, thus, can learn a new task while keeping the predictive ability on the earlier tasks. The architecture is built dynamically and starts with a single NN (named \textbf{column}). Equation~\ref{eq_pnn} shows that, for each new task $k$, a column is added whose \textit{i-th} layer receives the \textit{(i-1)-th} layer's output $h_{i-1}^{(k)}$ of column $k$ and the \textit{(i-1)-th} layers' outputs $h_{i-1}^{(j)}$ of all the earlier columns. $W_i$ and $U_i$ are the weight matrices to be learned. $U_i$ are called \textbf{lateral connections} and implement transfer learning.
\begin{equation}
\label{eq_pnn}
    h_i^{(k)} = f \left( W_i^{(k)} h_{i-1}^{(k)} + \sum_{j<k}^{} U_i^{(k:j)} h_{i-1}^{(j)}) \right)
\end{equation}

PNNs, on their own, do not tame temporal dependencies. To do so, they must use \textbf{RNNs} as columns, which operate on fixed-size sequences of items. RNNs recursively express the \textit{i-th} hidden layer's output $h_i$ as a function of the \textit{i-th} item's features $X_i$ and the output $h_{i-1}$ of the \textit{(i-1)-th} hidden layer. Due to the vanishing gradient, such an architecture cannot tame long temporal dependencies~\cite{q_deep_learning}. \textbf{Long short-term memory (LSTM)}~\cite{q_lstm} resolves this issue by memorizing only the helpful information and introducing the memory cell representing past cumulated knowledge. \textbf{Gated Incremental Memory (GIM)}~\cite{q_gim} develops a recurrent version of PNNs using LSTM as columns. Column $k$ receives lateral connections only from column $k-1$ to decrease the number of parameters. The \textit{i-th} item's hidden layer $h_i^{(k)}$ of column $k$ is computed as expressed by Equation \ref{eq_cpnn}. For each item $i$, its features $X_i$ and the previous column's \textit{i-th} hidden layer output are concatenated. Lateral connections are represented by the weights applied to the output of the previous column's hidden layer. The model's output is computed for each sequence element $i$ by applying a further layer after $h_i$.
\begin{equation}
\label{eq_cpnn}
    h_i^{(k)} = LSTM([X_i, h_i^{(k-1)}])
\end{equation}

The works mentioned above assume all data in each experience to be accessible at once. The specific paradigm called \textbf{Streaming Machine Learning (SML)}~\cite{book_bifet}, instead, was introduced to learn continually from a data point (or mini-batch) at a time (see Fig.~\ref{f_cl_sml}.b). \textbf{Concept drift}, that is a phenomenon in which the statistical properties of a target domain change over time in an arbitrary way~\cite{q_concept_drift}, is a crucial issue that SML tames.  We can distinguish two types of concept drift: virtual and real. It is easy to take them apart in the context of streaming classification. Virtual concept drifts do not affect the decision boundary, while real ones do. Additionally, in an abrupt drift, the new concept replaces the old one in a short period or in an exact instant, while in gradual and incremental drifts, the new concept gradually or incrementally replaces the old one. Finally, the concepts could reoccur over time. Concept drift detectors can detect all the mentioned types of concept drift~\cite{q_concept_drift}.

Most SML methods assume that the data stream's points are independent. In the real world, this assumption is unrealistic since they can exhibit temporal dependencies. Despite many works raising the issue that ignoring this situation can cause problems in the learning and evaluation processes~\cite{q_streaming_ts_read,Ziffer22,q_streaming_ts_zliobaite}, taming of temporal dependencies in an evolving data stream is still an open issue.

\section{Proposed Method}
\label{proposed}
This work proposes cPNN, a novel methodology for applying NNs to perform binary classification of an ESTS' data points. In Section~\ref{proposed_sgd_sml}, we analyze SGD behavior on data streams containing concept drifts. Section~\ref{proposed_sgd_ts} proposes a method to exploit SGD in an ESTS scenario. Finally, Section~\ref{proposed_cpnn} presents cPNN.

\subsection{Stochasting Gradient Descent for Evolving Data Streams}
\label{proposed_sgd_sml}
\begin{figure}
\centering
\includegraphics{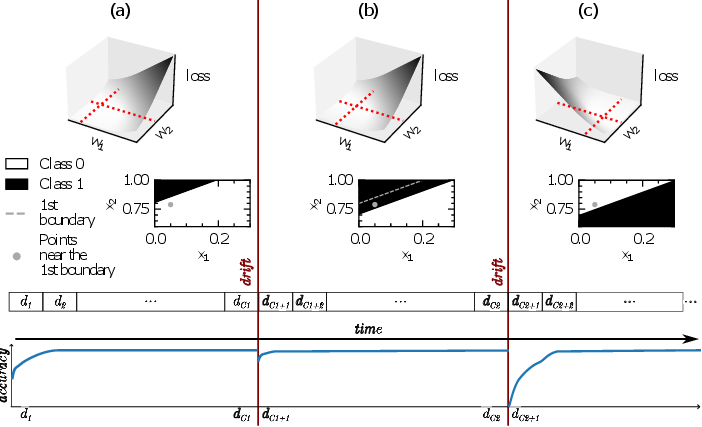}
\caption{Loss functions' minimum and accuracy trend of a single linear neuron associated with the following classification functions: (a) $-x_1+x_2-0.8\geq0$ (b) $-x_1+x_2-0.7\geq0$ (c) $-x_1+x_2-0.7<0$.} \label{f_min_loss}
\end{figure}

The SGD's iterative nature makes it possible to apply it on data streams by buffering the data points in fixed-size batches~\cite{q_deep_learning}.\footnote{See MOA's Perceptron application: \url{https://www.cs.waikato.ac.nz/~abifet/MOA/API/classmoa_1_1classifiers_1_1functions_1_1_perceptron.html}} Fig.~\ref{f_min_loss} illustrates this idea by analyzing a NN composed of a single linear neuron with two weights and no bias. Let's assume that the NN at $d_{C1}$, when a first abrupt drift occurs, has learned the decision boundary illustrated in Fig.~\ref{f_min_loss}.a. Notice that the second concept only marginally modifies the boundary between classes. Thus, SGD can quickly adapt to the drift since the minimum of the new concept's loss function is close to the previous one. On the contrary, the third concept swaps the classes when it occurs at $d_{C2}$. In this case, the new minimum is distant, and the SGD algorithm requires more iterations to reach it. Furthermore, the performance initially collapses since the starting configuration optimizes the inverted problem. In any case, when the model adapts to the new concept, it forgets the previous one since SGD has reached the new minimum. The more the new decision boundary changes, the lower the performance. Thus, a simple NN cannot deal with CF.

\subsection{Stochasting Gradient Descent for Streaming Time Series}
\label{proposed_sgd_ts}
\begin{figure}
\centering
\includegraphics{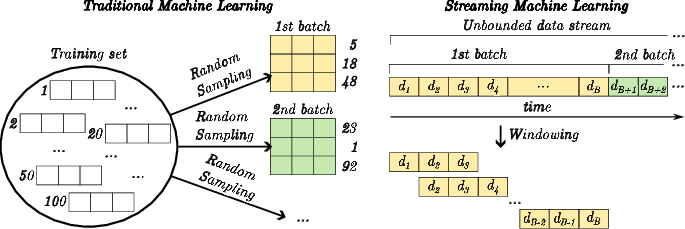}
\caption{Data processing in cases of Traditional Machine Learning and SML.} \label{f_windowing}
\end{figure}

As already stressed, although the i.i.d. assumption is usually made for each concept, data can show time dependence that requires RNN models like LSTM. To ensure that SGD is an unbiased gradient, we cannot sample an entire i.i.d. training set~\cite{q_sgd_shuffling_1}. The data points are, in fact, not available at once, and data has autocorrelations. We, therefore, input the data points in chronological order. Notice that, in this way, we are not minimizing the loss function to all the data but only to the most recently seen data points~\cite{q_importance_new_data}. Indeed, the literature on data streams~\cite{q_dsms} commonly assigns greater weight to recent data points because we expect that future data points related to the current concept will bear greater similarity to recent data. In particular, we adopt windowing from Data Stream Management Systems to propose (see Fig.~\ref{f_windowing}) to buffer data points in a batch with size B and build the sequences using a sliding window (with size W) once the batch is complete. In this way, we produce B-W+1 sequences for each batch. Notably, the windowing approach permits us to keep the temporal order.

\subsection{Continuous PNN (cPNN)}
\label{proposed_cpnn}
To better adapt to the concept drift, we propose a methodology to combine the knowledge gained from previous concepts with that learned from the current one. At the same time, we deal with catastrophic forgetting and, thus, provide accurate predictions for all the concepts. Moreover, we handle data points arriving continually from an unbounded data stream and tame temporal dependencies.

\begin{figure}
\centering
\includegraphics{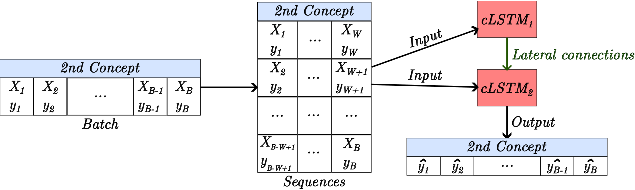}
\caption{The cPNN architecture during the second concept training.} \label{f_cpnn}
\end{figure}

PNNs and GIM can recycle old knowledge and avoid CF but are meant to be applied to CL experiences. We, thus, combine SML and CL techniques to build \textbf{Continuous PNN (cPNN)}: a continuous version of PNNs. We first define \textbf{Continuous LSTM (cLSTM)}, a continuous version of LSTM whose input is built as explained in Section~\ref{proposed_sgd_ts}. cLSTM outputs a probability distribution for each sequence item. Each data point's probability distribution on the target classes is computed by averaging its probability distributions associated with all the sequences to which it belongs. We then consider each concept as a task of CL. We use cLSTM as the base model (column) of cPNN to learn continually from an unbounded data stream's data points and tame temporal dependencies. Lateral connections are implemented as suggested by GIM to reduce the parameters. The architecture (shown in Fig.~\ref{f_cpnn}) can be edited by changing the column's type from cLSTM to any RNN model.

Algorithm~\ref{pseudo_model_training} details the cPNN's lifecycle. The architecture initially has a single column. We buffer the data stream in a batch with size B (Line \ref{pseudo_buffer}) and create the model's input (Line \ref{pseudo_input}) when the batch is complete. We, then, apply Prequential evaluation~\cite{q_prequential} (Lines \ref{pseudo_start_preq}-\ref{pseudo_end_preq}), taking first the model's predictions and evaluating the performance on the entire batch. Finally, we train the model on the batch for several epochs. After a concept drift, the model receives the batch accumulated up to that time (Line \ref{pseudo_batch_complete}). Then, we add a new column to the architecture, building lateral connections and freezing the weights of the previous column (Line \ref{pseudo_add_column}). Since CL assumes that the label associated with each experience is known and experiences are not mixed, we also assume that drifts are abrupt and to know when they occur. We rely on the presence of a concept label $c_t$ for each data point, which is the same for all the data points in a batch.
\begin{algorithm}[!ht]
\caption{\mbox cPNN training}
\label{pseudo_model_training}
\begin{algorithmic}[1]
\item[\textbf{Input:} Data stream \textit{S}, Batch size \textit{B}, Epochs \textit{E}, Window Size \textit{W}.]
\STATE \textit{batch} $\leftarrow$ \textit{empty list}, \textit{perf} $\leftarrow$ \textit{empty list}, $c_{t-1} \leftarrow -1$, $model \leftarrow$ \textit{new} $cPNN()$
\FORALL{$(X_t, y_t, c_t)$ in $S$}
    \STATE $drift \leftarrow True$
    \IF{$c_t = c_{t-1}$}
        \STATE Append $(X_t, y_t)$ to \textit{batch} \label{pseudo_buffer}
        \STATE $drift \leftarrow False$
    \ENDIF
    \IF{$|batch|=B$ OR $drift=True$} \label{pseudo_batch_complete}
        \IF{$|batch| \geq W$}
            \STATE $X,Y \leftarrow BuildSequences(batch, W)$ \label{pseudo_input}
            \STATE $pred \leftarrow model.predict(X)$ \label{pseudo_start_preq}
            \STATE Append $Evaluate(pred, Y)$ to $perf$
            \STATE $model.fit(X, Y, E)$ \label{pseudo_end_preq}
        \ENDIF
        \STATE \textit{batch} $\leftarrow$ \textit{empty list} \label{pseudo_clear}
    \ENDIF
    \IF{$drift=True$}
        \STATE $model.addColumn()$ \label{pseudo_add_column}
        \STATE Append $(X_t, y_t)$ to \textit{batch}
    \ENDIF
    \STATE $c_{t-1} \leftarrow c_t$
\ENDFOR
\end{algorithmic}
\end{algorithm}

\section{Experimental Evaluation}
This Section presents our experiments. Section~\ref{exp_set_gen} explains the generation of the data streams used in the ablation study described in Section~\ref{exp_set_ev}.
\label{exp_set}
\subsection{Generated Data Streams}
\label{exp_set_gen}
As detailed in~\cite{q_benchmark}, the most commonly used SML benchmarks containing temporal dependencies (Electricity and CoverType) are unsuitable for our purpose. The most known synthetic data stream generators (SINE~\cite{q_gama_learning_with_dd}, SEA~\cite{q_sea}, Hyperplane~\cite{q_hyperplane}, and STAGGER~\cite{q_stagger}) do not, instead, introduce temporal dependencies in the data. We, thus, propose the construction of a synthetic generator to have a simple and controlled case study to apply the models and analyze their behaviors. We start from SINE and produce a variant whose generated points have temporal dependencies. We begin from a randomly generated two-dimension point in (0,1). Each coordinate of the following points is generated by summing a random value (random walk~\cite{q_random_walk}) to the previously generated point's value. Every random walk's sign is generated to prevent exceeding the range (0,1). After quantifying the autocorrelation between data points using the Partial Autocorrelation Function plot, we set the maximum size of data points having the same label as ten. To identify the boundaries of the classes, we utilize the two SINE generator's \textbf{boundary functions} defined in Equation~\ref{eq_sine1} and \ref{eq_sine2}.
\begin{equation}
\label{eq_sine1}
S1: x_1 - sin(x_2) = 0 \quad \saclass{}
\end{equation}
\begin{equation}
\label{eq_sine2}
S2:x_1 - 0.5 - 0.3 \;sin(3\;\pi\;x_2)=0 \quad \sbclass{}
\end{equation}

We denote by \saplus{} and \sbplus{} the \textbf{classification functions} that classify with "1" the points above, respectively, the S1 and S2 curves, while with "0" the remaining ones. \saminus{} and \sbminus{} invert the labels of \saplus{} and \sbplus{} respectively. We generate one data stream for every classification function, each representing one concept and containing 50k data points. Let us introduce the term \textbf{sign drift} as the drift where a new concept reverses the labels while maintaining the boundary function (e.g. from \saplus{} to \saminus{}) or changing it (e.g., from \saplus{} to \sbminus{}). We combine the data streams in two ways. Firstly, \textbf{classification inversion drift} produces a single sign drift that keeps the boundary function unchanged (Fig.~\ref{f_min_loss}.c). Secondly, \textbf{boundary function drift} combines all four concepts' data streams by alternating the boundary functions (Fig.~\ref{f_min_loss}.b) and producing one or two sign drifts. By design, more than 50\% of the sequences with the same label have a maximum length of five, and labels are balanced. When we change the boundary function without a sign drift (e.g., from \saplus{} to \sbplus{}), 65\% of the points keep the same label. If we combine a sign drift with a boundary function drift (e.g., from \saplus{} to \sbminus{}), the percentage drops to 35\%. Finally, all the points change their labels after a classification inversion drift (e.g., from \saplus{} to \saminus{}).

\subsection{Experimental Setting}
\label{exp_set_ev}
We conduct an ablation study for our hypothesis formulation and compare cPNN with two alternative architectures. \textbf{mcLSTM} (Multiple cLSTMs) remove the lateral connections so that each new column does not consider the previous column's hidden layer output. Direct application of the base model \textbf{cLSTM} (see Section~\ref{proposed_cpnn}) removes, instead, the creation of different columns, resulting in a cPNN with only one column that ignores drifts. Hyperparameter values are chosen as follows after executing the preliminary experiments. Epochs number: 10, window size: 10, batch size: 128, learning rate: 0.1, hidden's layer size: 50.\footnote{Complete source code available at \url{https://github.com/federicogiannini13/cpnn}} The final performance is computed by averaging the batches. Since the labels are balanced and we do not focus on a particular class, we evaluate the accuracy.

Our hypothesis is that cPNN can adapt to new concepts in an ESTS more quickly than the other two architectures. Additionally, we expect that models can quickly adapt to a new concept if it is similar to the previous one. A sign drift would be more complicated if the model does not learn to invert its past knowledge. 
We evaluate the final accuracy in four \textbf{cases} to verify the two hypotheses for each concept. The first two cases (\textbf{[1,50]} and \textbf{[1,100]}) analyze how models adapt to the new concept by considering the accuracy at the end of the first 50 and 100 batches after the drift.\footnote{Thresholds represent the number of batches during preliminary experiments, after which the more robust and less robust models achieved good performance.} A reasonably accurate model in the first part of the concept is robust to concept drifts. The third case (\textbf{[100,)}), which covers the batch range from 100 onwards, assesses the accuracy of models in response to the newly introduced concept once they have adapted to it. Finally, the fourth case (\textbf{[1,)}) monitors the entire concept by investigating accuracy from the first batch onwards. Each experiment is repeated ten times, and their average accuracy is analyzed. Tables~\ref{t_class_inv_drift}, \ref{t_boundary_function_drift_1} and \ref{t_boundary_function_drift_2} of Section~\ref{results} report results using the mentioned notations.

\section{Results}
\label{results}
This section analyzes the results of the different experiments described in Section~\ref{exp_set_ev}. Tables~\ref{t_class_inv_drift}, \ref{t_boundary_function_drift_1}  and \ref{t_boundary_function_drift_2} report the ten executions' average accuracies and standard deviations. Since the first concept's architectures are the same, we make comparisons from the second concept onwards. Architectures are compared in pairs. We report in bold the statistically best-performing architecture (if it is statistically better performing than the remaining two) and in italics the less-performing one. We, thus, first conduct a Shapiro-Wilk test to check for normality. If we cannot reject the null hypothesis for both distributions, we conduct a Welch's t-test. Otherwise, we run a Wilcoxon signed-rank test. We perform a one-sided test in both cases. We underline the not normally distributed samples. All the tests are conducted with a significance level of 0.05.

\subsection{Classification Inversion Drift}
\label{results_sign}
Results in Table~\ref{t_class_inv_drift} show that after the concept drift, cLSTM performance collapse. Since they are similar, we only report results for two data streams. For the S1 classification function, the mcLSTM's random initialization of the parameters works better than the cLSTM one (which is the inverse concept's optimal one), but from the 100th till the end of the concept, cLSTM outperforms mcLSTM. cPNN can adapt quickly to the new concept. It results in being the best-performing model in all the experiments. In the case of S2, the gap between cPNN and the other models is more significant. These experiments suggest that cPNN could learn to invert past knowledge. cLSTM requires more iterations to reach the new optimal setting since it starts from the inverse concept one. At the end of each concept, cLSTM's new optimal configuration is still worse than cPNN's one.

\begin{table}[!t]\caption{Accuracies on classification inversion drift. cPNN outperforms the ablated versions in all cases.}\label{t_class_inv_drift}
\centering
\resizebox{\textwidth}{!}{
\begin{tabular}{rr|ccc|l|ccc|}
\cline{3-5} \cline{7-9} 
\multicolumn{1}{c} {\textbf{}}               &                  & \multicolumn{3}{c|} { \textbf{\saplus{} \saminus{}}}                                                                                                            &  & \multicolumn{3}{c|}{\textbf{\sbplus{} \sbminus{}}}                                                                                                            \\ \cline{1-5} \cline{7-9} 
\multicolumn{1}{|r|} {\textbf{concept}}      & \textbf{case} & \multicolumn{1}{c|}{\textbf{cPNN}}       & \multicolumn{1}{c|}{\textbf{cLSTM}}                             & \textbf{mcLSTM}                            &  & \multicolumn{1}{c|}{\textbf{cPNN}}       & \multicolumn{1}{c|}{\textbf{cLSTM}}                             & \textbf{mcLSTM}                            \\ \cline{1-5} \cline{7-9} 
\multicolumn{1}{|r|}{}                      & {[}1,50{]}       & \multicolumn{1}{c|}{\textbf{.96, .004}}  & \multicolumn{1}{c|}{{ \textit{.872, .026}}} & .903, .008                                 &  & \multicolumn{1}{c|}{{\ul \textbf{.864, .017}}} & \multicolumn{1}{c|}{{ .742, .024}} & .75, .006                                  \\
\multicolumn{1}{|r|}{}                      & {[}1,100{]}      & \multicolumn{1}{c|}{\textbf{.972, .002}} & \multicolumn{1}{c|}{{ \textit{.921, .014}}} & .933, .004                                 &  & \multicolumn{1}{c|}{{\ul \textbf{.888, .02}}}  & \multicolumn{1}{c|}{{ .79, .024}}           & { \textit{.774, .006}} \\
\multicolumn{1}{|r|}{}                      & (100,)           & \multicolumn{1}{c|}{\textbf{.989, .001}} & \multicolumn{1}{c|}{.98, .002}                                  & { \textit{.975, .001}} &  & \multicolumn{1}{c|}{{\ul \textbf{.927, .018}}} & \multicolumn{1}{c|}{.883, .029}                                 & { \textit{.834, .009}} \\
\multicolumn{1}{|r|}{\multirow{-4}{*}{2nd}} & {[}1,)           & \multicolumn{1}{c|}{\textbf{.984, .001}} & \multicolumn{1}{c|}{.965, .004}                                 & .964, .001                                 &  & \multicolumn{1}{c|}{{\ul \textbf{.917, .018}}} & \multicolumn{1}{c|}{.859, .026}                                 & { \textit{.818, .007}} \\ \cline{1-5} \cline{7-9} 
\end{tabular}
}
\end{table}

\subsection{Boundary Function Drift}
\label{results_boundary}
Results regarding boundary function drift (shown in Tables~\ref{t_boundary_function_drift_1} and ~\ref{t_boundary_function_drift_2}) indicate that cPNN adapts more quickly to a new concept after a sign drift and when the new boundary function is more complex than the previous (a drift from S1 to S2). In this case, cPNN outperforms the other architectures in the first 50 and 100 batches. From the 100th batch, cLSTM and cPNN have similar performance. cLSTM outperforms cPNN in the first batches only after the first drift from S2 to S1, with no sign drift. mcLSTM performs worse in almost all the experiments.
 \begin{table}[!t]\caption{Accuracies on data streams \saplus{}, \sbplus{}, \saminus{}, \sbminus{} and \saplus{}, \sbminus{}, \saminus{}, \sbplus{}. cPNN always recovers faster from concept drifts than the ablated versions. In some cases, a single cLSTM performs better in the long run, but in the end, it only remembers that last concept since it does not manage CF. mcLSTM that does not use transfer learning and resets the parameter configuration performs worse in almost all situations.}\label{t_boundary_function_drift_1}
\centering
\resizebox{\textwidth}{!}{
\begin{tabular}{rr|ccc|l|ccc|}
\cline{3-5} \cline{7-9}
\multicolumn{1}{c}{\textbf{}}               &                  & \multicolumn{3}{c|}{\textbf{\saplus{} \sbplus{} \saminus{} \sbminus{}}}                                                                                       &                                                       & \multicolumn{3}{c|}{\textbf{\saplus{} \sbminus{} \saminus{} \sbplus{}}}                                                                                                        \\ \cline{1-5} \cline{7-9} 
\multicolumn{1}{|r|}{\textbf{concept}}      & \textbf{case} & \multicolumn{1}{c|}{\textbf{cPNN}}             & \multicolumn{1}{c|}{\textbf{cLSTM}}            & \textbf{mcLSTM}                                  &                                                       & \multicolumn{1}{c|}{\textbf{cPNN}}             & \multicolumn{1}{c|}{\textbf{cLSTM}}                             & \textbf{mcLSTM}                                  \\ \cline{1-5} \cline{7-9} 
\multicolumn{1}{|r|}{}                      & {[}1,50{]}       & \multicolumn{1}{c|}{{\ul \textbf{.759, .012}}} & \multicolumn{1}{c|}{.746, .011}                & { \textit{.738, .006}}       &                                                       & \multicolumn{1}{c|}{\textbf{.781, .006}} & \multicolumn{1}{c|}{{ \textit{.743, .015}}} & .753, .006                                       \\
\multicolumn{1}{|r|}{}                      & {[}1,100{]}      & \multicolumn{1}{c|}{\textbf{.803, .007}}       & \multicolumn{1}{c|}{.791, .012}                & { \textit{.762, .005}}       &                                                       & \multicolumn{1}{c|}{{\ul \textbf{.808, .004}}}       & \multicolumn{1}{c|}{.772, .008}                                 & .775, .004                                       \\
\multicolumn{1}{|r|}{}                      & (100,)           & \multicolumn{1}{c|}{.877, .006}                & \multicolumn{1}{c|}{ \textbf{.897, .013}} & { \textit{.829, .011}}       &                                                       & \multicolumn{1}{c|}{\textbf{.876, .005}}       & \multicolumn{1}{c|}{.851, .011}                           & { \textit{.823, .009}}       \\
\multicolumn{1}{|r|}{\multirow{-4}{*}{2nd}} & {[}1,)           & \multicolumn{1}{c|}{.858, .005}                & \multicolumn{1}{c|}{ \textbf{.87, .012}}  & { \textit{.812, .009}}       &                                                       & \multicolumn{1}{c|}{\textbf{.859, .003}}       & \multicolumn{1}{c|}{.831, .009}                           & { \textit{.811, .007}}       \\ \cline{1-5} \cline{7-9} 
\multicolumn{1}{|r|}{}                      & {[}1,50{]}       & \multicolumn{1}{c|}{\textbf{.951, .004}}       & \multicolumn{1}{c|}{.893, .022}                & .905, .008                                       &                                                       & \multicolumn{1}{c|}{\textbf{.947, .004}}       & \multicolumn{1}{c|}{.94, .009}                                  & { \textit{.908, .004}}       \\
\multicolumn{1}{|r|}{}                      & {[}1,100{]}      & \multicolumn{1}{c|}{\textbf{.964, .004}}       & \multicolumn{1}{c|}{.932, .013}       & .935, .004                                       &                                                       & \multicolumn{1}{c|}{\textbf{.961, .003}}       & \multicolumn{1}{c|}{.955, .006}                                 & { \textit{.936, .003}}       \\
\multicolumn{1}{|r|}{}                      & (100,)           & \multicolumn{1}{c|}{.982, .002}                & \multicolumn{1}{c|}{{\ul .982, .001}}          & { \textit{.975, .001}}       & { \textit{}}                      & \multicolumn{1}{c|}{.981, .002}                & \multicolumn{1}{c|}{ .982, .001}                           & { \textit{.975, .001}}       \\
\multicolumn{1}{|r|}{\multirow{-4}{*}{3rd}} & {[}1,)           & \multicolumn{1}{c|}{\textbf{.977, .002}}       & \multicolumn{1}{c|}{.969, .004}                & { \textit{.965, .001}}       &                                                       & \multicolumn{1}{c|}{.976, .002}                & \multicolumn{1}{c|}{.975, .002}                                 & { \textit{.965, .001}}       \\ \cline{1-5} \cline{7-9} 
\multicolumn{1}{|r|}{}                      & {[}1,50{]}       & \multicolumn{1}{c|}{\textbf{.855, .012}}       & \multicolumn{1}{c|}{.778, .016}                & { {\ul \textit{.754, .005}}} &                                                       & \multicolumn{1}{c|}{\textbf{.862, .008}}       & \multicolumn{1}{c|}{.777, .025}                                 & {  \textit{.738, .005}} \\
\multicolumn{1}{|r|}{}                      & {[}1,100{]}      & \multicolumn{1}{c|}{\textbf{.88, .011}}        & \multicolumn{1}{c|}{.822, .012}                & { \textit{.776, .004}}       & \multicolumn{1}{c|}{{ \textit{}}} & \multicolumn{1}{c|}{{\ul\textbf{.88, .006}}}        & \multicolumn{1}{c|}{{\ul .831, .019}}                                 & { \textit{.76, .003}}        \\
\multicolumn{1}{|r|}{}                      & (100,)           & \multicolumn{1}{c|}{{\ul .907, .007}}          & \multicolumn{1}{c|}{.91, .009}                 & { \textit{.826, .01}}        & \multicolumn{1}{c|}{{ \textit{}}} & \multicolumn{1}{c|}{.909, .009}         & \multicolumn{1}{c|}{.915, .013}                                 & { \textit{.827, .009}}       \\
\multicolumn{1}{|r|}{\multirow{-4}{*}{4th}} & {[}1,)           & \multicolumn{1}{c|}{{\ul \textbf{.9, .008}}}   & \multicolumn{1}{c|}{.888, .009}                & \textit{.813, .008}                                       & \multicolumn{1}{c|}{}                                 & \multicolumn{1}{c|}{.902, .008}          & \multicolumn{1}{c|}{{\ul.894, .014}}                                 & { \textit{.81, .007}}        \\ \cline{1-5} \cline{7-9} 
\end{tabular}
}

\end{table}
\begin{table}[!t]\caption{Accuracies on data streams \sbplus{}, \saplus{}, \sbminus{}, \saminus{} and \sbplus{}, \saminus{}, \sbminus{}, \saplus{}.}\label{t_boundary_function_drift_2}
\centering
\resizebox{\textwidth}{!}{
\begin{tabular}{rr|ccc|l|ccc|}
\cline{3-5} \cline{7-9}
\multicolumn{1}{c}{\textbf{}}               &                  & \multicolumn{3}{c|}{\textbf{\sbplus{} \saplus{} \sbminus{} \saminus{}}}                                                                                            &                                & \multicolumn{3}{c|}{\textbf{\sbplus{} \saminus{} \sbminus{} \saplus{}}}                                                                                                        \\ \cline{1-5} \cline{7-9} 
\multicolumn{1}{|r|}{\textbf{concept}}      & \textbf{case} & \multicolumn{1}{c|}{\textbf{cPNN}}                              & \multicolumn{1}{c|}{\textbf{cLSTM}}      & \textbf{mcLSTM}                            &                                & \multicolumn{1}{c|}{\textbf{cPNN}}                                    & \multicolumn{1}{c|}{\textbf{cLSTM}}            & \textbf{mcLSTM}                            \\ \cline{1-5} \cline{7-9} 
\multicolumn{1}{|r|}{}                      & {[}1,50{]}       & \multicolumn{1}{c|}{{\ul .931, .014}}                           & \multicolumn{1}{c|}{.933, .012}          & { \textit{.915, .004}} &                                & \multicolumn{1}{c|}{{\ul \textbf{.921, .008}}}                        & \multicolumn{1}{c|}{.892, .022}                & .903, .003                                 \\
\multicolumn{1}{|r|}{}                      & {[}1,100{]}      & \multicolumn{1}{c|}{{\ul .943, .01}}                            & \multicolumn{1}{c|}{\textbf{.951, .008}} & .941, .003                                 &                                & \multicolumn{1}{c|}{{\ul .939, .009}}                                 & \multicolumn{1}{c|}{.934, .012}                & .934, .003                                 \\
\multicolumn{1}{|r|}{}                      & (100,)           & \multicolumn{1}{c|}{{ \textit{.974, .003}}} & \multicolumn{1}{c|}{\textbf{.983, .001}} & .976, .001                                 &                                & \multicolumn{1}{c|}{{ {\ul \textit{.971, .004}}}} & \multicolumn{1}{c|}{{\ul \textbf{.983, .001}}} & .975, .001                                 \\
\multicolumn{1}{|r|}{\multirow{-4}{*}{2nd}} & {[}1,)           & \multicolumn{1}{c|}{.966, .005}                                 & \multicolumn{1}{c|}{\textbf{.974, .003}} & .967, .001                                 &                                & \multicolumn{1}{c|}{{\ul .963, .005}}                                 & \multicolumn{1}{c|}{\textbf{.97, .003}}        & .964, .001                                 \\ \cline{1-5} \cline{7-9} 
\multicolumn{1}{|r|}{}                      & {[}1,50{]}       & \multicolumn{1}{c|}{\textbf{.824, .013}}                        & \multicolumn{1}{c|}{.768, .023}          & .754, .005                                 &                                & \multicolumn{1}{c|}{\textbf{.835, .013}}                              & \multicolumn{1}{c|}{.779, .016}                & { \textit{.755, .007}} \\
\multicolumn{1}{|r|}{}                      & {[}1,100{]}      & \multicolumn{1}{c|}{\textbf{.851, .015}}                        & \multicolumn{1}{c|}{.802, .024} & { \textit{.774, .004}} &                                & \multicolumn{1}{c|}{\textbf{.863, .01}}                               & \multicolumn{1}{c|}{.812, .018}                & { \textit{.776, .003}} \\
\multicolumn{1}{|r|}{}                      & (100,)           & \multicolumn{1}{c|}{.896, .01}                                  & \multicolumn{1}{c|}{.892, .018}          & { \textit{.835, .011}} & \textit{}                      & \multicolumn{1}{c|}{.903, .009}                                       & \multicolumn{1}{c|}{.893, .017}                & { \textit{.831, .01}}  \\
\multicolumn{1}{|r|}{\multirow{-4}{*}{3rd}} & {[}1,)           & \multicolumn{1}{c|}{\textbf{.885, .011}}                        & \multicolumn{1}{c|}{.869, .02}           & { \textit{.819, .009}} &                                & \multicolumn{1}{c|}{\textbf{.893, .009}}                              & \multicolumn{1}{c|}{.872, .017}                & { \textit{.817, .008}} \\ \cline{1-5} \cline{7-9} 
\multicolumn{1}{|r|}{}                      & {[}1,50{]}       & \multicolumn{1}{c|}{\textbf{.952, .006}}                        & \multicolumn{1}{c|}{.942, .007}          & { \textit{.907, .008}} &                                & \multicolumn{1}{c|}{\textbf{.953, .007}}                              & \multicolumn{1}{c|}{.923, .017}                & .92, .004                                  \\
\multicolumn{1}{|r|}{}                      & {[}1,100{]}      & \multicolumn{1}{c|}{\textbf{.963, .004}}                        & \multicolumn{1}{c|}{.957, .003}          & { \textit{.936, .005}} & \multicolumn{1}{c|}{\textit{}} & \multicolumn{1}{c|}{\textbf{.962, .004}}                              & \multicolumn{1}{c|}{.945, .01}                 & {\ul .944, .002}                           \\
\multicolumn{1}{|r|}{}                      & (100,)           & \multicolumn{1}{c|}{.98, .004}                                  & \multicolumn{1}{c|}{\textbf{.983, .002}} & { \textit{.975, .001}} & \multicolumn{1}{c|}{\textit{}} & \multicolumn{1}{c|}{.979, .003}                                       & \multicolumn{1}{c|}{\textbf{.982, .002}}       & { \textit{.976, .001}} \\
\multicolumn{1}{|r|}{\multirow{-4}{*}{4th}} & {[}1,)           & \multicolumn{1}{c|}{.975, .004}                                 & \multicolumn{1}{c|}{.976, .001}          & { \textit{.965, .002}} & \multicolumn{1}{c|}{}          & \multicolumn{1}{c|}{.975, .003}                                       & \multicolumn{1}{c|}{.973, .004}                & { \textit{.967, .0}}   \\ \cline{1-5} \cline{7-9} 
\end{tabular}
}
\end{table}

\section{Conclusion}
\label{conclusion}
This paper pioneers a novel continuous version of PNNs for Evolving Streaming Time Series. We proposed CPNN to deal simultaneously with concept drifts and temporal dependencies while avoiding catastrophic forgetting. To do so, we presented a continuous adaptation of LSTM (namely cLSTM) that exploits the SGD algorithm to tame temporal dependencies in a data stream. A similar method was used by \cite{q_ilstm} on a complex architecture and real datasets. Instead, our goal was to analyze the models' behaviors using a simplified scenario. cPNN's architecture is based on PNNs to tame CF and use transfer learning to fit new concepts quickly. To investigate cPNN behavior, we generated synthetic data streams and conducted an ablation study. cPNN performance highlighted a quicker adaptation to new concepts. Its average accuracy after each concept drift is, in fact, statistically greater than the ablated ones. cPNN resulted, thus, in being more robust to concept drifts, especially in the case of sign drift. 

One of the main limitations of cPNN is that its complexity increases linearly with the number of concepts. We, thus, imagine that this architecture could be applied in the case of reoccurrent drifts where we would need to check whether the new concept has been seen before. Additionally, when dealing with data streams, the selection of hyperparameters can become challenging, and the resulting outcomes may be highly sensitive to these choices. Moreover, we only studied the models in a simplified scenario with abrupt concept drifts and synthetic data streams containing only two features. In our future works, we intend to explore more types of drift in a higher dimensional space and complex classification functions. Finally, as in many CL experiments, we assumed to have an "oracle" that knows the concept associated with each data point. In our future works, we will apply concept drift detection methods. cPNN performance results suggested that it could automatically learn to invert past knowledge when there is a sign drift. We also think its quicker adaptation to the new concept is due to past recycling ability. We will analyze the model's parameters in future works to verify it. In the long term, we intend to investigate how cPNN learns in contexts where real data evolves via gradual or incremental concept drifts. We will most likely need to examine other types of columns, like Gated Recurrent Units or Transformers.

\bibliographystyle{splncs04}

\bibliography{bibliography}

\end{document}